\documentclass[letterpaper, 10 pt, conference]{ieeeconf}  

\IEEEoverridecommandlockouts                              

\usepackage{graphics,caption} 
\graphicspath{ {figure/} }
\usepackage{graphicx} 
\usepackage{epsfig} 
\usepackage{times} 
\usepackage{amsmath} 
\usepackage{amssymb} 
\usepackage{cite}
\usepackage{booktabs}
\usepackage{threeparttable}
\usepackage{multirow}
\if CLASSOPTIONcompsoc
\usepackage[caption=false, font=normalsize, labelfont=sf, textfont=sf]{subfig}
\else
\usepackage[caption=false, font=footnotesize]{subfig}
\usepackage{url}
\usepackage{mathrsfs}
\usepackage[colorlinks,linkcolor=blue]{hyperref}

\usepackage{float}
\usepackage{stfloats}

\usepackage{tcolorbox}
\tcbuselibrary{skins}
\usepackage{lipsum}

\pdfoutput=1

\title{\LARGE \bf
PANav: Toward Privacy-Aware Robot Navigation \\
via Vision-Language Models
}

\author{Bangguo Yu, Hamidreza Kasaei, Ming Cao
\thanks{*This work of Yu is supported in part by the China Scholarship Council.}
\thanks{All authors are with the Faculty of Science and Engineering, University of Groningen, 9747 AG Groningen, the Netherlands. {\tt\small \{b.yu, hamidreza.kasaei, m.cao\}@rug.nl}}%
}

\begin{document}

\maketitle
\thispagestyle{empty}
\pagestyle{empty}

\begin{abstract}
Navigating robots discreetly in human work environments while considering the possible privacy implications of robotic tasks presents significant challenges. Such scenarios are increasingly common, for instance, when robots transport sensitive objects that demand high levels of privacy in spaces crowded with human activities. While extensive research has been conducted on robotic path planning and social awareness, current robotic systems still lack the functionality of privacy-aware navigation in public environments.
To address this, we propose a new framework for mobile robot navigation that leverages vision-language models to incorporate privacy awareness into adaptive path planning. Specifically, all potential paths from the starting point to the destination are generated using the A* algorithm. Concurrently, the vision-language model is used to infer the optimal path for privacy-awareness, given the environmental layout and the navigational instruction. This approach aims to minimize the robot's exposure to human activities and preserve the privacy of the robot and its surroundings.
Experimental results on the S3DIS dataset demonstrate that our framework significantly enhances mobile robots' privacy awareness of navigation in human-shared public environments. Furthermore, we demonstrate the practical applicability of our framework by successfully navigating a robotic platform through real-world office environments. 
The supplementary video and code can be accessed via the following link: \href{https://sites.google.com/view/privacy-aware-nav}{https://sites.google.com/view/privacy-aware-nav}.

\end{abstract}

\section{INTRODUCTION}

\begin{figure}[htbp]
    \centering
    \includegraphics[scale=0.65]{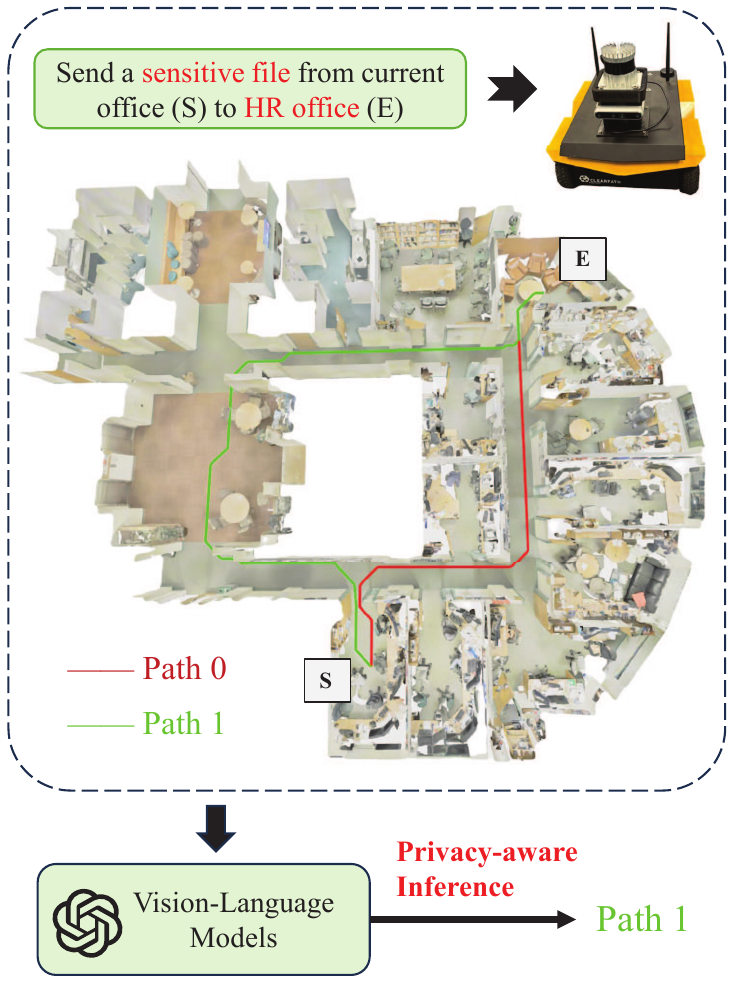}
    \caption{Privacy-aware navigation example. The robot uses vision language models to find a more private path (shown in green) based on the environment map and the navigational instructions.}
    \label{fig:demo}
    \vspace{-0.5cm}
\end{figure}

Privacy is a critical concern in robotics, particularly when considering interactive and autonomous robots operating in public environments. 
These robots often collect vast amounts of personal and sensitive data about users and their surroundings, making privacy a central topic not only for static datasets but also for sensitive dynamic interactions.
While recent advancements in robot perception and planning have greatly enhanced robots' ability to navigate complex, unstructured environments, privacy considerations remain underdeveloped—especially in scenarios involving the transport of sensitive objects in public spaces. Consider the scenario where a robot is tasked with delivering a classified document from a manager's office to the HR department. There are multiple paths the robot could take, and if it simply follows the shortest route, it might pass through crowded office areas or pedestrian-heavy corridors. To ensure the safety and privacy of the navigation, the robot must be equipped with sufficient scene understanding and privacy awareness to generate a path that minimizes the exposure of sensitive documents in populated areas, prioritizing privacy and discretion over mere efficiency.

Navigation in known environments is a classical task in robotics, typically involving the planning of the shortest path from the start to the destination using an environment map. The last few years have witnessed the development of path planning \cite{Hart1968}\cite{Dijkstra1959}\cite{LaValle2001}, scene understanding \cite{Gu2023}\cite{Feng2023a}, and socially aware navigation \cite{Chen2019b}\cite{Chen2017a}\cite{Song2024}, cumulatively leading to huge progress in navigation tasks within human work environments. Traditional methods for robot navigation rely on geometric building maps for path planning, which struggle to understand the semantic information of the environment.
The rise of machine learning has transformed this landscape by enabling learning-based perception models that enhance scene understanding and human activity recognition, garnering significant attention and achieving notable success in different navigation tasks. Most socially aware navigation frameworks focus on human-robot interactions, such as maintaining a comfortable distance from humans \cite{Chen2017a}\cite{Jin2020a} or mimicking human decision-making based on observed trajectories \cite{Chen2019b}. However, these approaches typically use only limited information from the specific path and fail to incorporate a comprehensive understanding of the entire environment.
With the advent of large pre-trained models, a promising approach is to leverage large language models or vision-language models to transfer powerful knowledge priors into navigation tasks. For instance, approaches \cite{Long2023}\cite{Long2024}\cite{Driess2023} use large pre-trained models to perceive the world and make decisions regarding actions. Similarly, approaches \cite{Yu2023c}\cite{Chen2024}\cite{Song2024} utilize these models to gain a global understanding of the environment and make informed navigational decisions. However, while these methods effectively exploit the perception and planning capabilities of large models, they have yet to address the hidden needs and influences of human participants fully.
What kind of navigation task is best suited for large models that can think more like humans? This motivates us to explore how vision-language models can be harnessed to incorporate privacy-aware decision-making into navigation tasks.

This paper addresses the challenge of privacy-aware robot navigation tasks, wherein the robot needs to transport sensitive objects in public environments by selecting the optimal path not only in terms of distance but also considering privacy preservation. Specifically, we present PANav, a novel framework that leverages vision-language models (VLMs) to develop an adaptive path planner for privacy-aware navigation, ensuring the privacy of the robot and potentially its surroundings. An illustrative example of privacy-aware robot navigation for transporting a sensitive file is shown in Fig. \ref{fig:demo}. After analyzing the environment map and identifying two possible routes from the current office to the HR department, the VLM selects the more private option (\emph{Path 1}), minimizing exposure in heavily populated office areas based on the layout of the public environment.
Our framework is evaluated using the S3DIS dataset \cite{Armeni2016a}, which includes various public office environments, and is compared against classical path-planning methods. Unlike many other navigation approaches, PANav utilizes vision-language models to infer privacy concerns, meaning that the true needs and influence of humans are considered during navigation. This leads to significant improvements in protecting the privacy of both the robot and human activities in the environment. Additionally, we demonstrate the applicability of our framework in real-world scenarios. To the best of our knowledge, this is the first work to introduce privacy-aware robot navigation using vision-language models.

Our contributions are summarized as follows:

\begin{itemize}
    \item We propose a novel privacy-aware robot navigation framework that can be executed for navigation in public environments, ensuring the safety and privacy of the robot and its surroundings. 
    \item  We make the first attempt to integrate top-view maps of the environment into the decision-making process of vision-language models, enabling adaptive path planning based on privacy awareness and demonstrating the effectiveness of large models in social navigation tasks.
    \item  We validate our method on the S3DIS dataset and apply it to a real robot equipped with an RGB-D camera and a LiDAR sensor in an office environment, demonstrating the method’s effectiveness and practical applicability.
    \item  We propose a new evaluation metric using a Gaussian-modulated distance field for privacy-aware navigation tasks. 
\end{itemize}

\section{RELATED WORK}

\subsection{Privacy Aware Robotics}

Privacy is an essential topic in social robotics and becomes even more critical when considering interactive and autonomous robots in public environments. Previous work has proposed various types of privacy in robotics, such as conversations with robots \cite{Tang2022}, sensor selection for robots \cite{Eick2020}, motion planning \cite{Luo2020}, and other data or software safety concerns \cite{Heuer2021}. Specifically, the most important privacy considerations in public environments are the information about the robot itself and human activities. On one hand, the robot may unintentionally expose sensitive information while navigating crowded human environments. On the other hand, robots often collect a large amount of personal and sensitive data about users and their surroundings through cameras and other sensors. To address these challenges, we propose an adaptive path planner based on privacy awareness, leveraging vision-language models to protect the privacy of both the robot and human activities in public environments.

\subsection{Navigation in crowded environments}

Navigation in crowded environments has been a critical task for mobile robots in various scenarios, such as transportation in hotels or hospitals. Previous work has proposed various models that often utilize an environmental representation to guide the process and consider aspects like collision avoidance and the needs of bystanders. Specifically, understanding environmental information better relies on the representation of the map, which is crucial for the agent. Examples include grid maps \cite{Grisetti2005}\cite{Hess2016}, topological maps \cite{Esparza2013}\cite{Armeni}, semantic maps \cite{Gu2023}\cite{Grinvald2019a}, and implicit maps \cite{Mildenhall2022}\cite{Kerbl2023}.
Furthermore, navigation approaches mainly consist of planning-based and learning-based methods. Planning-based approaches are frequently used to make global-level decisions and plan the initial path for the robot to follow. These include search-based approaches such as A* \cite{Hart1968}, D* \cite{Stentz1995}, and Dijkstra's algorithm \cite{Dijkstra1959}, as well as sampling-based approaches like PRM \cite{Kavraki1996}, RRT \cite{LaValle2001}, and Fast Marching methods \cite{Sethian1999}.
Learning-based methods involve training a model from data to predict future states or decisions \cite{Chen2017a}\cite{Jin2020a}\cite{Chen2019b}. These methods can leverage large amounts of data to learn patterns and adapt to new situations, though they may require significant training time and might not generalize well to novel scenarios. Deep reinforcement learning, deep learning, and inverse reinforcement learning are commonly used for these tasks. They can help provide a good initial estimate to assist in better planning \cite{Chen2017a} or in selecting sub-goals to guide a local planner \cite{Brito2021}.

While large pre-trained models have achieved significant success, they can also be leveraged to understand the global environment and make decisions. VLMaps \cite{Huang2023} employs the LSeg \cite{Li2022} model to encode RGB images, aligning pixel embeddings with 3D map locations to create a semantic map, which is useful for semantic navigation. ConceptGraph \cite{Gu2023} proposes an open-vocabulary 3D scene graph representation that assigns features exclusively to object nodes, which can be applied in robot navigation. MapGPT \cite{Chen2024} utilizes large language models (LLMs) to interpret map information and achieve adaptive path planning. Co-NavGPT \cite{Yu2023c} similarly parses the environment using LLMs, assigning paths to multiple robots. VLM-SocialNav \cite{Song2024} employs a perception model to detect key social entities and prompts a VLM to generate guidance for socially compliant robot behavior. However, these methods have yet to address the privacy concerns of mobile robots in public human work environments. In contrast, our framework leverages privacy awareness from vision-language models, enabling a more private and secure navigation process compared to previous approaches.

\section{METHOD}
In this section, we describe the definition of privacy-aware navigation and the framework of leveraging vision-language models to plan the path based on privacy awareness.

\begin{figure*}[htbp]
    \centering
    \includegraphics[scale=0.52]{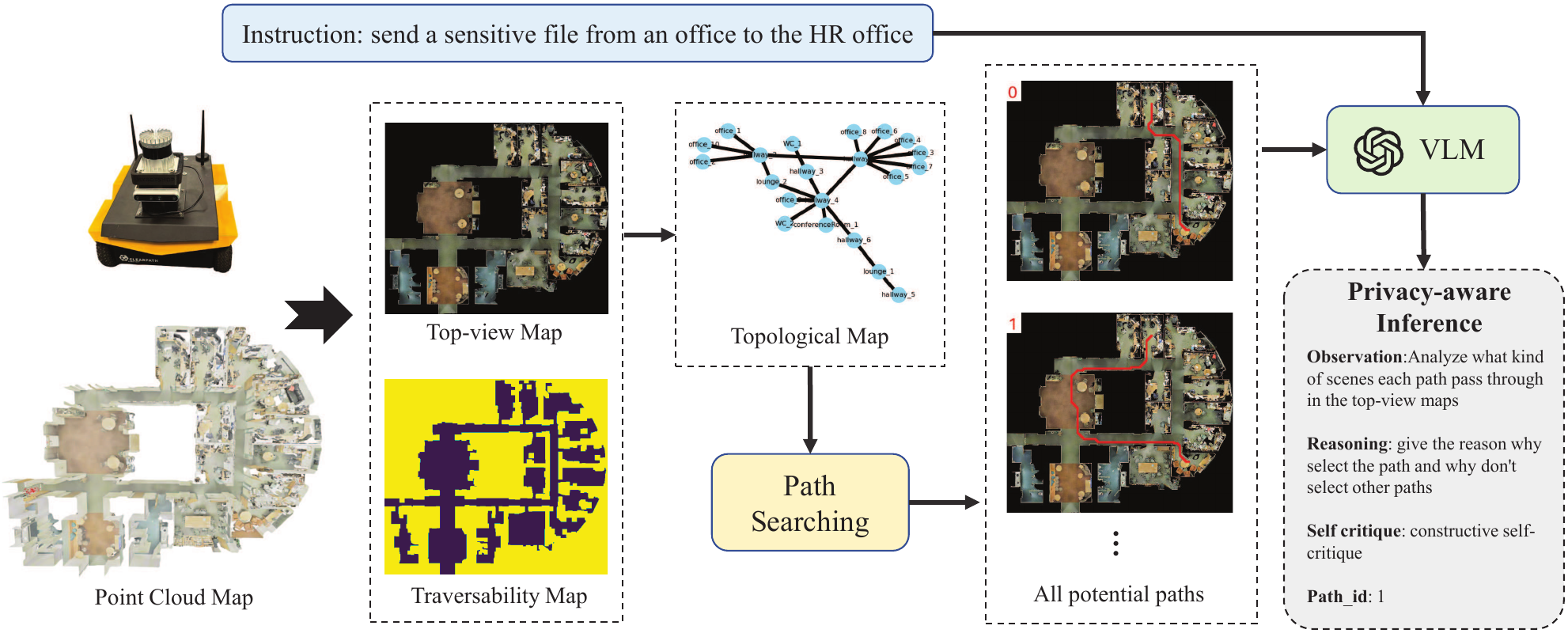}
    \caption{The architecture of the privacy-aware navigation framework. The framework takes a point cloud map as input to generate top-view and topological maps, and an optimal path is selected from all potential paths based on the privacy-aware inference of the vision-language models. Once the optimal path is selected, the robot can follow the path to complete the navigation task.}
    \label{fig:system_architecture}
    \vspace{-0.4cm}
\end{figure*}
 
\subsection{Task Definition}

Privacy-aware navigation involves the agent navigating a known environment with the adaptive path planner following the navigational instructions. The set of navigational instructions is represented by $D = \left\{d_0, \dots, d_m\right\}$, and the scenes are represented by $S = \left\{s_0, \dots, s_n\right\}$.
Each episode begins with the agent being initialized at a start position $p^s_i$ in the scene $s_i$ and receiving the navigational instruction $d_i$ and the destination $p^e_i$. Thus, an episode can be denoted as $T_i = \left\{s_i, d_i, p^s_i, p^e_i\right\}$. In each episode, the agent observes the environment point cloud and the instruction, and a private path should be generated from the starting point to the destination based on privacy awareness.

\subsection{Overview}

Our framework is illustrated in Fig. \ref{fig:system_architecture}. First, the agent obtains the point cloud of the scene to build the top-view map and traversability map. The topological map is also generated from the top-view and traversability maps. Second, all potential topological paths are extracted from the topological map, and the metric paths are generated by A* \cite{Hart1968} from start to destination. Then, vision-language models are used to infer the potential privacy areas of the environment, and an optimal path is selected from all the possible metric paths. After obtaining the optimal path, the agent can follow the path and navigate to the destination.

\subsection{Map Representation}
\subsubsection{Metric Map}

We construct the top-view map and traversability map from the point cloud of the scene. To get the top-view map $M_{top}$, we extract all the environment point clouds without the ceilings and project them into a 2D grid map. We only select the highest point of the point cloud for each grid cell to ensure that more environmental information can be recorded. For the traversability map $M_{tra}$, we extract all the grid cells that contain points within a high range of the point cloud as traversable cells, and all the grids not occupied by the top-view map are set as obstacle regions. In this way, a clearer traversability map without any walls is built. The coordinate system and resolution of the top-view and traversability maps are aligned to maintain the same scale.

\subsubsection{Topological Map}

Since we want to find all possible paths from the start to the destination rather than only the shortest one, the topological map is constructed from the annotated point cloud. We separate all types of annotated scenes with different point cloud sets and calculate the minimum distances between them. Next, we build a topology $G_T = <O, E>$ for the environment: each scene in the point cloud is represented as a node $o_i \in O$ in the graph, such as \emph{office\_1}, \emph{hallways\_2}, and we only find the edges $e_{i,j} \in E$ between the hallways node $o_i$ and other scenes $o_j$ based on the minimum distance of the point clouds, which allows all the rooms in the environment to be connected via hallways. We also align the centers of all nodes with the traversability map, which can then be used for path generation.

\subsection{Path Generation}

After we have obtained a topological connectivity graph representing the structure of the environment, all the possible paths from the start to the destination can be determined. First, all the simple paths are extracted from the topological map. A simple path is a path between two nodes in a graph that does not revisit any node. In other words, a simple path contains no cycles. We use the depth-first search (DFS) \cite{TARJANR1971} strategy to explore all simple paths between two nodes, and each path is a list containing all the scenes the path passes through.
Next, we filter the paths based on the scenes they pass through and their length. Since we only want paths that go through the hallways on the map, any other scenes, such as offices or lobbies in the path, are removed. If one path in the set of paths is a subpath of another, the longer path is also removed to avoid meaningless loops. Then, the top-$k$ shortest topological paths $P_t = \left\{p^t_0, \dots, p^t_k\right\}$ are selected as candidate paths for further path selection, provided the total number of simple paths exceeds $k$. 

After getting all possible topological paths $P_t$, the metric paths $P_m = \left\{p^m_0, \dots, p^m_k\right\}$ can be generated based on the center position of each node $o_i$ aligned with the metric map. For each topological path $p^t_i$, A* \cite{Hart1968} is used to generate the optimal grid path $p^m_i$ between each pair of neighboring nodes on the traversability map $M_{tra}$, which can then be connected into a complete path from the start to the destination.
\begin{align}
    &P_t = \text{DFS}(G_T, p^s, p^e) \\
    &P_m = \text{A*}(P_t, M_{tra})
\end{align}
Thus, all the candidate potential paths $P_m$ in the metric maps can be obtained.

\subsection{Optimal Path Inference}

With all the information gathered and provided by previous modules, the agent needs to synthesize and reason over the current observed top-view map $M_{top}$, the navigational instruction $d$, and the set of all candidate paths $M_p$. The instruction contains the object description with the start node $p^s$ and the target node $p^e$ in the topological map $G_T$, such as \emph{Send a classified file from the manager's office to the HR department}. A strong path selection module is required to leverage all the information effectively.

While designing such a module from scratch is nearly infeasible, we utilize powerful vision-language models directly as the path selection module with designed prompts to reason over all the information and select the optimal path. In particular, we separately add each candidate path $p^m_i$ into the top-view map $M_{top}$ with the identified path ID $i$ to obtain a candidate path list with maps $M_p = \left\{m_{p_0}, \dots, m_{p_k}\right\} $, Then, we input all the candidates $M_p$ into the VLMs along with the navigational instruction $d$. The candidate paths $M_p$ can be seen in Fig. \ref{fig:system_architecture}. The VLMs will infer the optimal path from all the candidates based on privacy awareness and output the selected path ID. This formalization makes it easier for the VLMs to make decisions without requiring any few-shot demonstrations. Furthermore, to obtain the optimal path $p^*$, we perform five independent runs of the VLMs to generate a candidate path set $\left\{ p_i \right\}_{i=1}^5$. The most frequently occurring path in this set is selected as the optimal path.

In summary, the proposed PANav, which combines the candidate map list with paths $M_p$, navigational instruction $d$, the start position $p^s$, and the destination $p^e$, can be defined as follows:
\begin{equation}
    p^* = \arg\max_{m_p \in M_p} \sum_{i=1}^{5} \mathbb{I}(p = \text{VLMs}_i(M_p, d, p^s, p^e))
\end{equation}
Where $p^*$ is the selected path from $M_p$.
We also use the zero-shot chain-of-thought prompting technique introduced by \cite{Kojima2022} to encourage the VLMs to perform more reasoning before giving the final answer.

\section{EXPERIMENTS}

In this section, we evaluate the performance of our method by comparing it with the classical path planner method using the 3D reconstruction scene dataset. Additionally, we apply our method on a real-world robot platform to validate its practicality for navigational tasks.

\subsection{Simulation Experiment}

\renewcommand\arraystretch{1.4}

\begin{table}
    \centering
    \caption{Navigational Instructions}
    \setlength{\tabcolsep}{1mm}{}
    {
        \begin{tabular}{cc}
            \toprule
            Scenario & Navigational Instructions                                  \\
            \midrule
            1 & send a classified file from the office to the HR office          \\
            2 & send fragile equipment from the office to the meeting room  \\
            3 & send medicine from the office to the bathroom             \\
            \bottomrule
        \end{tabular}
    }
    \label{tab:instruction}
    \vspace{-0.4cm}
\end{table}

\begin{table*}[htbp]
    \centering
    \fontsize{9}{8}\selectfont
    \begin{threeparttable}
        \caption{Results of Comparative Study in S3DIS.}
        \label{tab:comparation_study}
        \setlength{\tabcolsep}{4mm}{}
        {
            \begin{tabular}{ccccccc}
                \toprule
                \multirow{2}{*}{Method}  & \multicolumn{2}{c}{Area\_3} & \multicolumn{2}{c}{Area\_4} & \multicolumn{2}{c}{Area\_5a} \\
                \cmidrule(lr){2-3} \cmidrule(lr){4-5} \cmidrule(lr){6-7}
                                    & $P_{risk}$    & Distance      & $P_{risk}$    & Distance      & $P_{risk}$    & Distance \\
                \midrule
                A*\cite{Hart1968}   & 1928.649      & 3373          & 682.212       & 4470          & 2390.677      & 5884 \\
                Ours                & 960.534       & 4973          & 321.152       & 7220          & 1564.794      & 8880 \\
                \bottomrule
            \end{tabular}
        }
        
    \end{threeparttable}
    \vspace{-0.1cm}
\end{table*}

\begin{figure*}[htbp]
\centering
\subfloat[path\_0]
{
    \includegraphics[width=3.2cm]{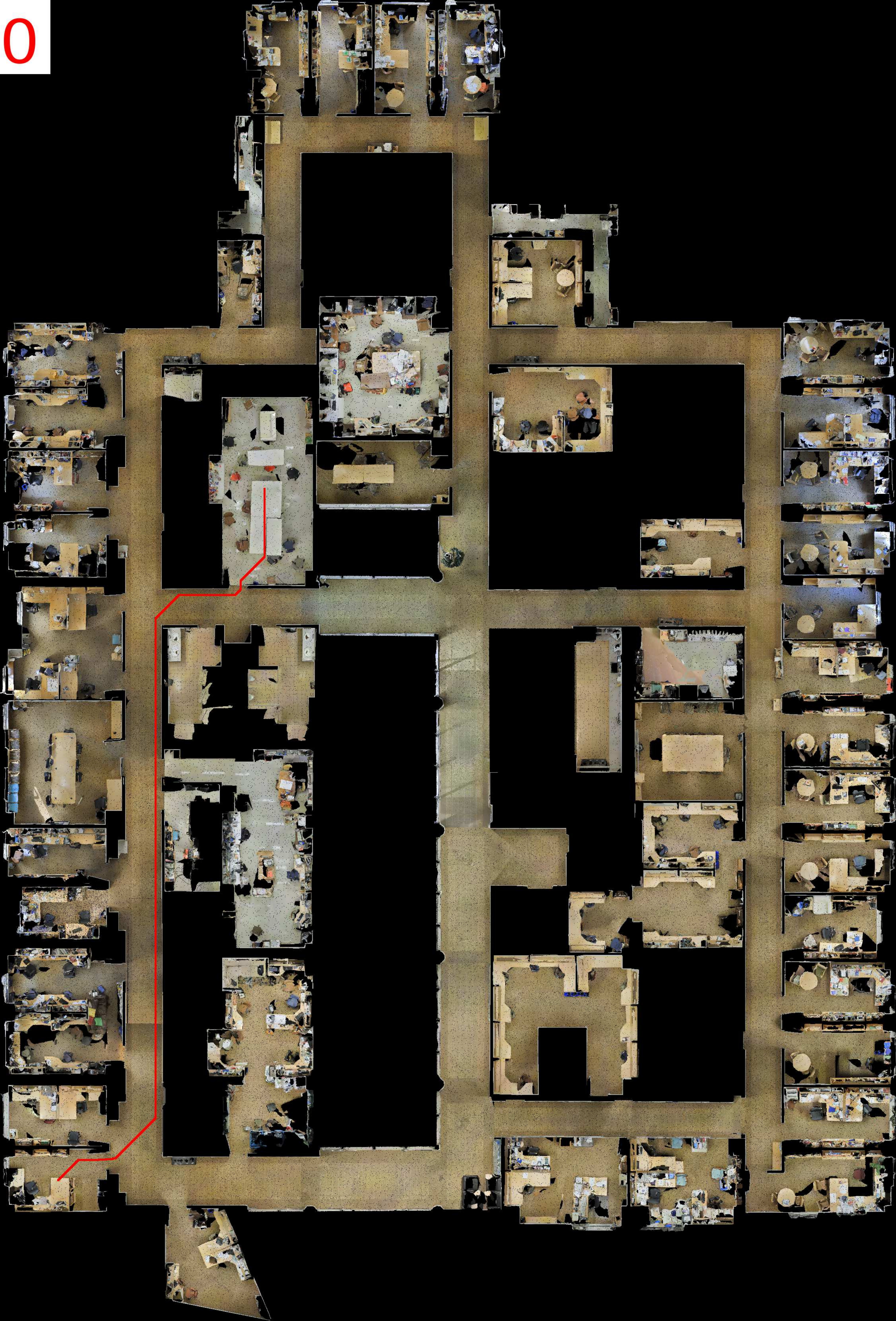}
}
\subfloat[path\_1]
{
    \includegraphics[width=3.2cm]{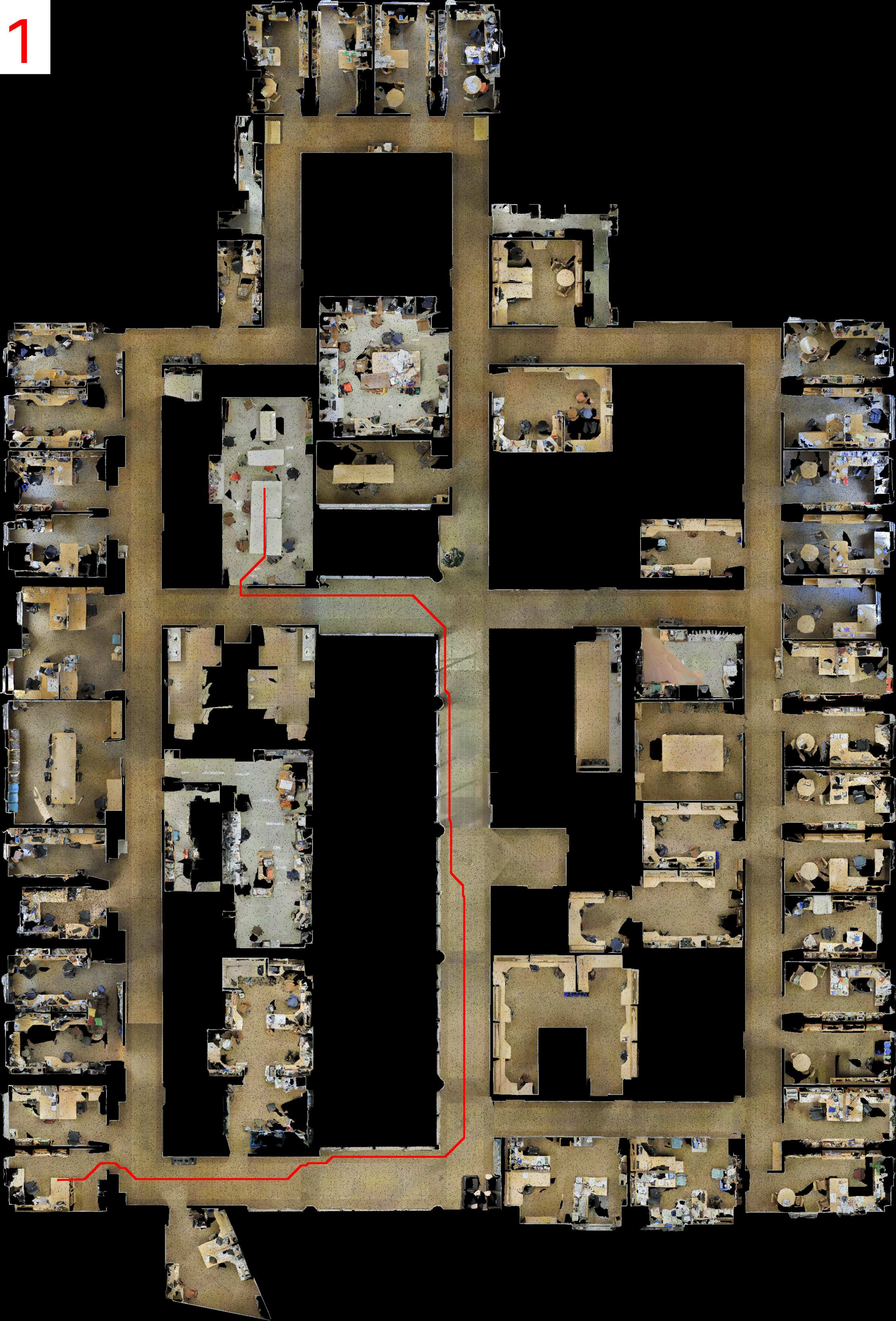}
}
\subfloat[path\_2]
{
    \includegraphics[width=3.2cm]{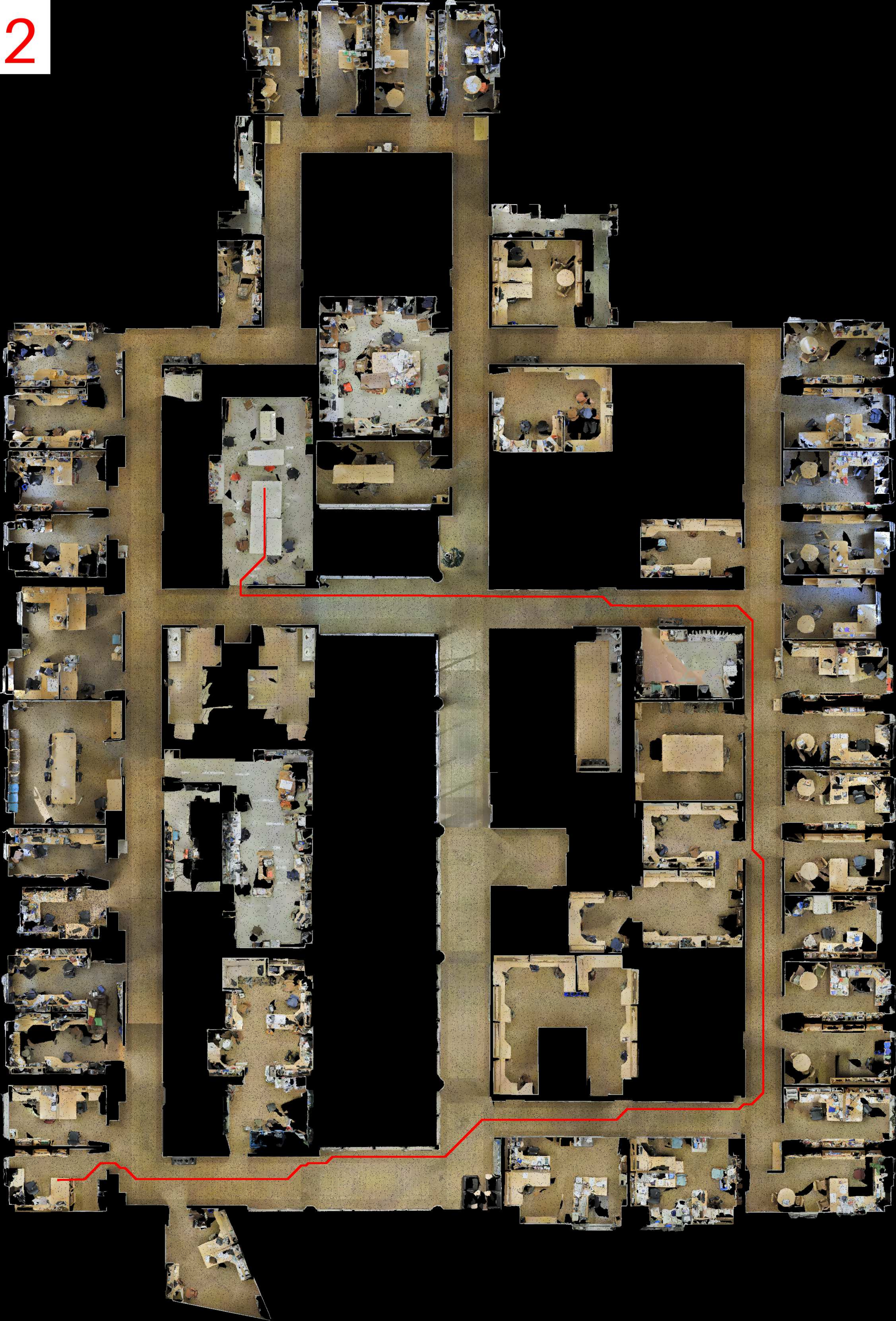}
}
\subfloat[path\_3]
{
    \includegraphics[width=3.2cm]{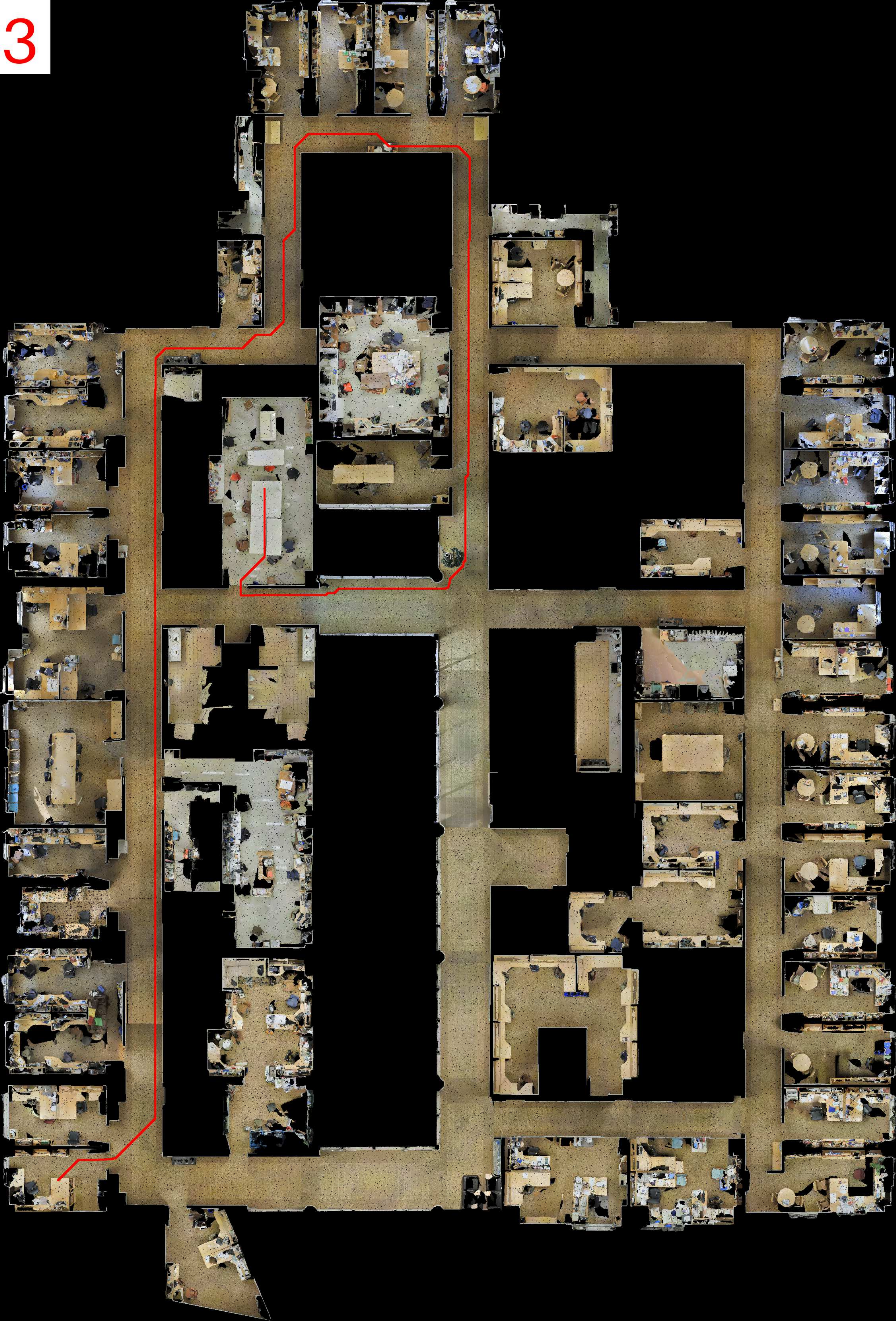}
}
\subfloat[path 4]
{
    \includegraphics[width=3.2cm]{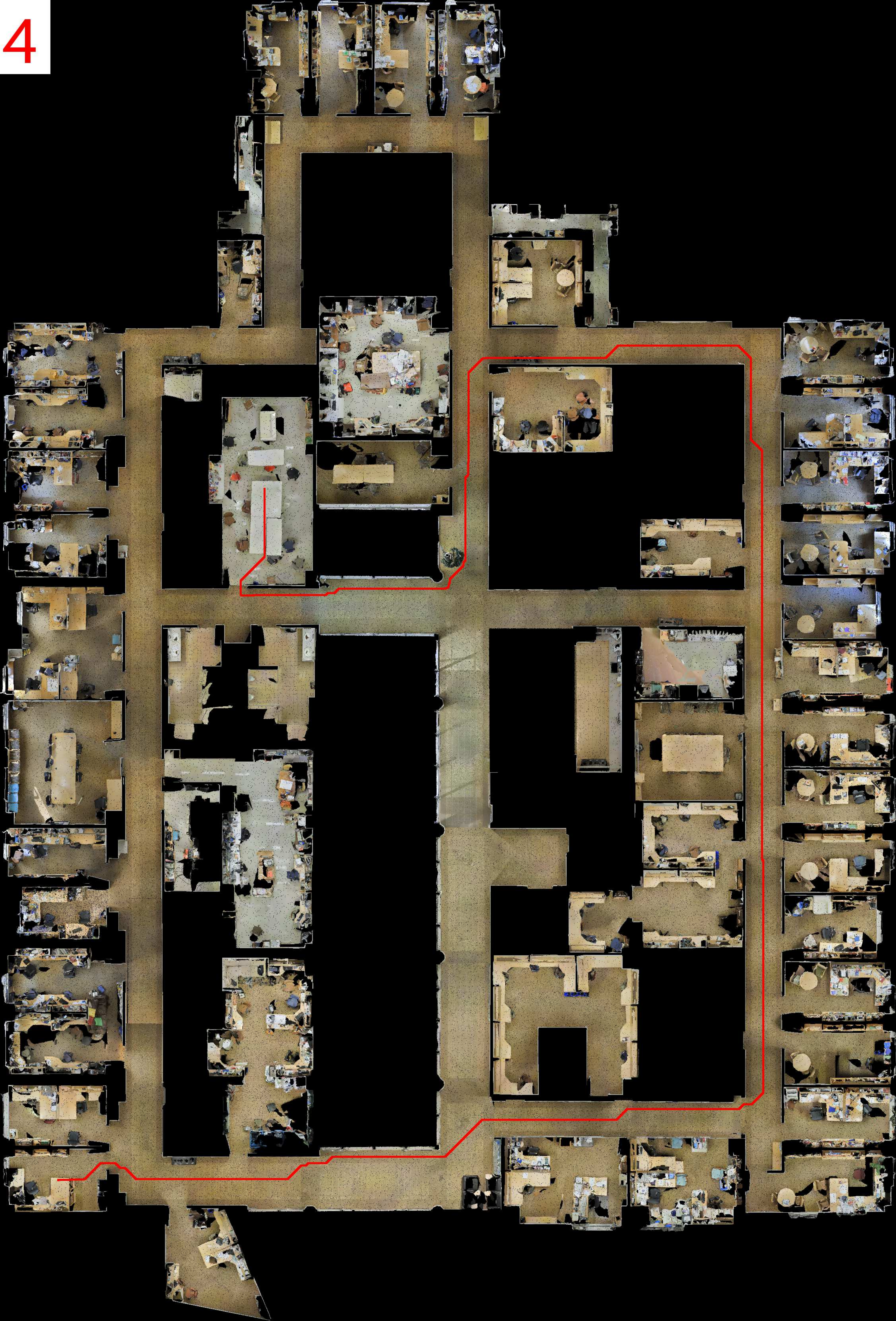}
}
\caption{A case of the top-5 possible paths for transporting a classified file from an office to a conference room in \emph{Area\_5a}. The red lines represent the paths extracted from the topological map.}
\label{fig:sim_path}
\vspace{-0.4cm}
\end{figure*}

\subsubsection{Dataset}

Our experiments are conducted on the S3DIS dataset \cite{Armeni2016a}, which provides a variety of mutually registered modalities from 3D domains, with instance-level semantic and geometric annotations. It covers over 6,000 $m^2$ collected in 6 large-scale indoor areas originating from 3 different buildings. It also includes registered raw and semantically annotated 3D meshes and point clouds. We selected 3 buildings: \emph{Area\_3}, \emph{Area\_4}, and \emph{Area\_5a} from the dataset as the experimental scenes. For each building, we construct three episodes, with each episode selecting a pair of start and destination rooms from the annotated rooms to evaluate privacy awareness. Each episode starts in an office, and the destinations are the office, conference room, and bathroom, respectively. We also prepared the related navigational instructions for each episode, which are shown in TABLE\ref{tab:instruction}.

\subsubsection{Experiment Setting}

Since we only needed to find the optimal path from the start to the destination, we conducted our evaluation on the point clouds from the S3DIS dataset without using any simulation platform. GPT-4 Turbo \cite{openai2023gpt4turbo} was used as the VLM to select the path from all candidate paths, and the temperature parameter was set to 0.5. To evaluate the navigation performance of our framework, we considered the classical path planner A*\cite{Hart1968} as the baseline. The A* algorithm is a heuristic function-based algorithm for efficient path planning. It calculates the heuristic function's value at each node in the work area and involves checking many adjacent nodes to find the optimal solution, with zero probability of collision. 

\subsubsection{Evaluation Metrics}

\begin{figure}[htbp]
    \centering
    \includegraphics[scale=0.1]{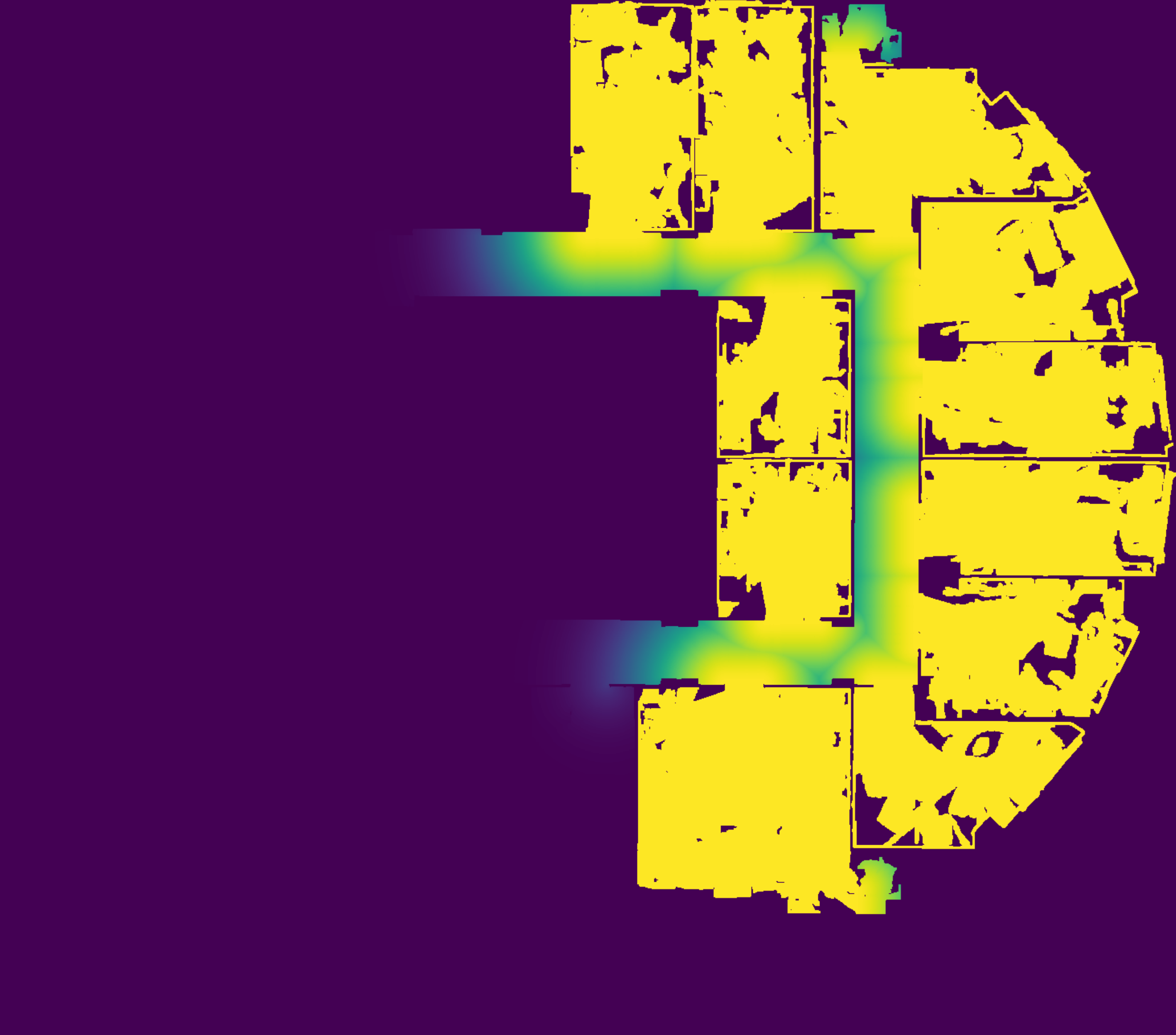}
    \caption{Gaussian-modulated distance field based on the traversability map.}
    \label{fig:evaulation_map}
    \vspace{-0.4cm}
\end{figure}

To evaluate the privacy of each path in the environment, we generate a Gaussian-modulated distance field based on the traversability map for each scene, as shown in Fig. \ref{fig:evaulation_map}. It takes into account specific areas (such as "office" or "conference" rooms) and manipulates the traversability maps to produce a traversable field. We mask all the map grids in the office or conference room areas as obstacles and calculate the distance field for other traversable areas relative to the obstacle areas in the traversability map. The distance field $D(x, y)$ is calculated using the Fast Marching Method (FMM) \cite{Sethian1999}, which gives the shortest distance from each traversable cell to the nearest obstacle:
\begin{equation}
    D(x, y) = \text{FMM}(M_{tra}(x, y))
\end{equation}
Gaussian modulation creates a smooth, continuous field that favors regions farthest from obstacles. It transforms the distance field into a scalar field, prioritizing paths based on the distance to obstacles in the traversability map.
\begin{equation}
    G(x, y) = \exp\left( -\frac{(D(x, y) - \mu)^2}{2\sigma^2} \right)
\end{equation}
where \( G(x, y) \) represents the value of the Gaussian-modulated field at the point \( (x, y) \), \( \mu = \max(D(x, y)) \) is the maximum value of the distance field, \( \sigma \) is the standard deviation, typically set as \( \sigma = \frac{\mu}{\sigma_d} \), where \( \sigma_d \) is a scaling factor.
This field is used as an evaluation map, where higher values correspond to more crowded areas for the robot to navigate. We then obtain all point values along the path in the distance field, using the sum of these values as the privacy-aware risk score $P_{risk}$ for each path $p$:
\begin{equation}
    P_{risk}(p) = \sum_{(x, y) \in p} G(x, y)
\end{equation}
Higher scores along the path indicate a greater risk of exposing privacy. The distance of each path is also used to evaluate the navigation performance.

\subsubsection{Result and Discussion}

The results summarized in Table \ref{tab:comparation_study}  reveal the difference in the performance of the two methods. Our method consistently achieved a lower risk of exposing privacy across all tested areas, indicating a superior capability in maintaining privacy. Specifically, In \emph{Area\_3}, our approach achieved a $P_{risk}$ of 960.534 with a path distance of 4,973 units, whereas the A* algorithm had a higher $P_{risk}$ of 1,928.649 and a shorter distance of 3,373 units. Similar patterns emerged in \emph{Area\_4} and \emph{Area\_5a}, where our method consistently lowered the $P_{risk}$ to 321.152 and 1,804.972, respectively, compared to the A* algorithm's scores of 682.212 and 2,390.677. These findings suggest that our framework effectively prioritizes privacy by selecting paths that minimize exposure to sensitive areas, as represented by the Gaussian-modulated distance field. The increased path distances highlight that our method prioritizes privacy over the shortest route, opting for paths that, while longer, significantly reduce the risk of privacy breaches. Fig. \ref{fig:sim_path} shows a case of all possible paths from an office to a conference room in \emph{Area\_5a}, where the selected path by our framework is \emph{path\_1} even the shortest is \emph{path\_0}, as \emph{path\_1} avoids high-traffic office areas and takes a route that appears to have less human activity. Our method's ability to flexibly weigh privacy concerns against travel efficiency makes it well-suited for scenarios where privacy is a priority, such as in environments that handle sensitive information or require discretion.


\subsection{Real World Experiment}

\begin{figure}[htbp]
    \centering
    \includegraphics[scale=0.10]{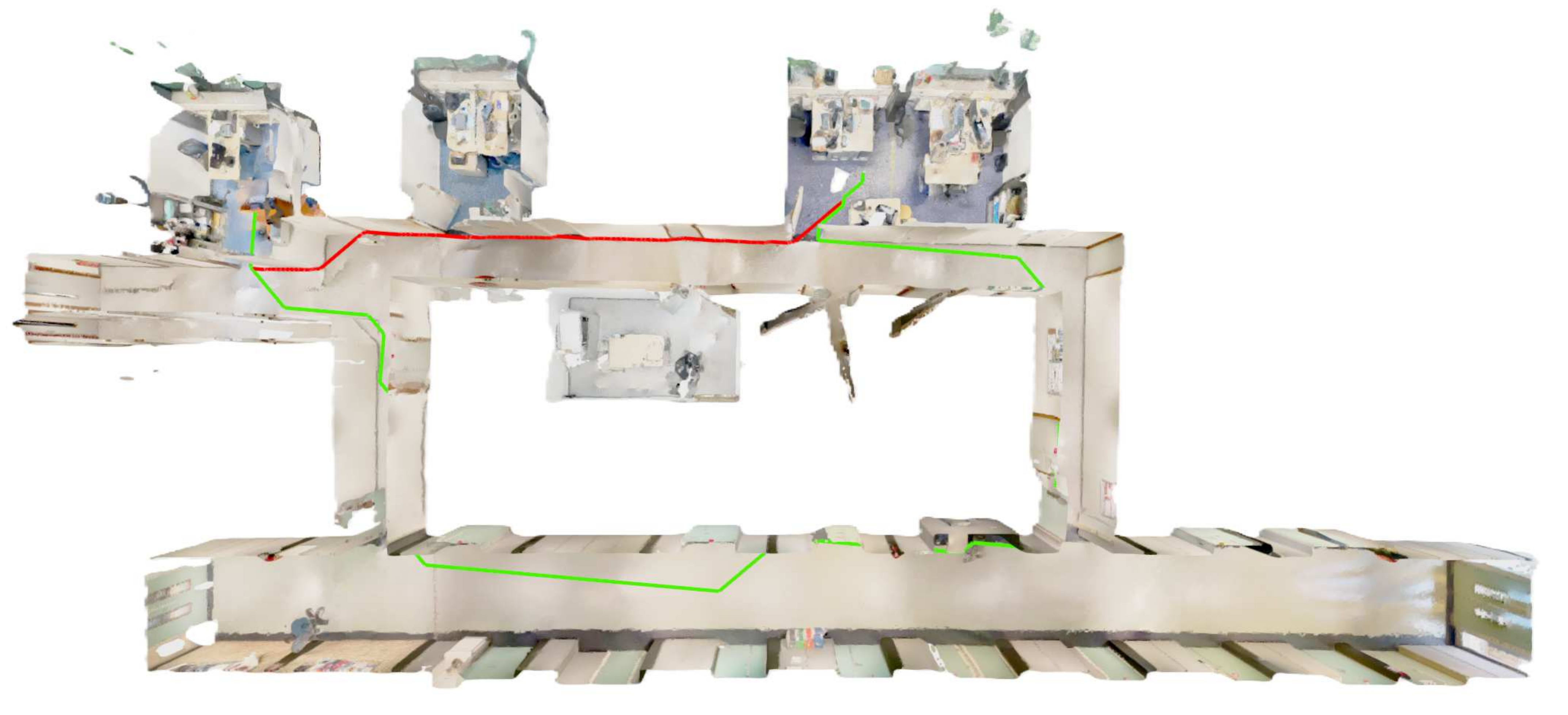}
    \caption{The 3D scan point cloud of the real office scene includes three office rooms, a conference room, and several corridors.}
    \label{fig:realscene}
\end{figure}

\begin{figure}[htbp]
\centering
\subfloat[path\_0]
{
    \includegraphics[width=2.585cm]{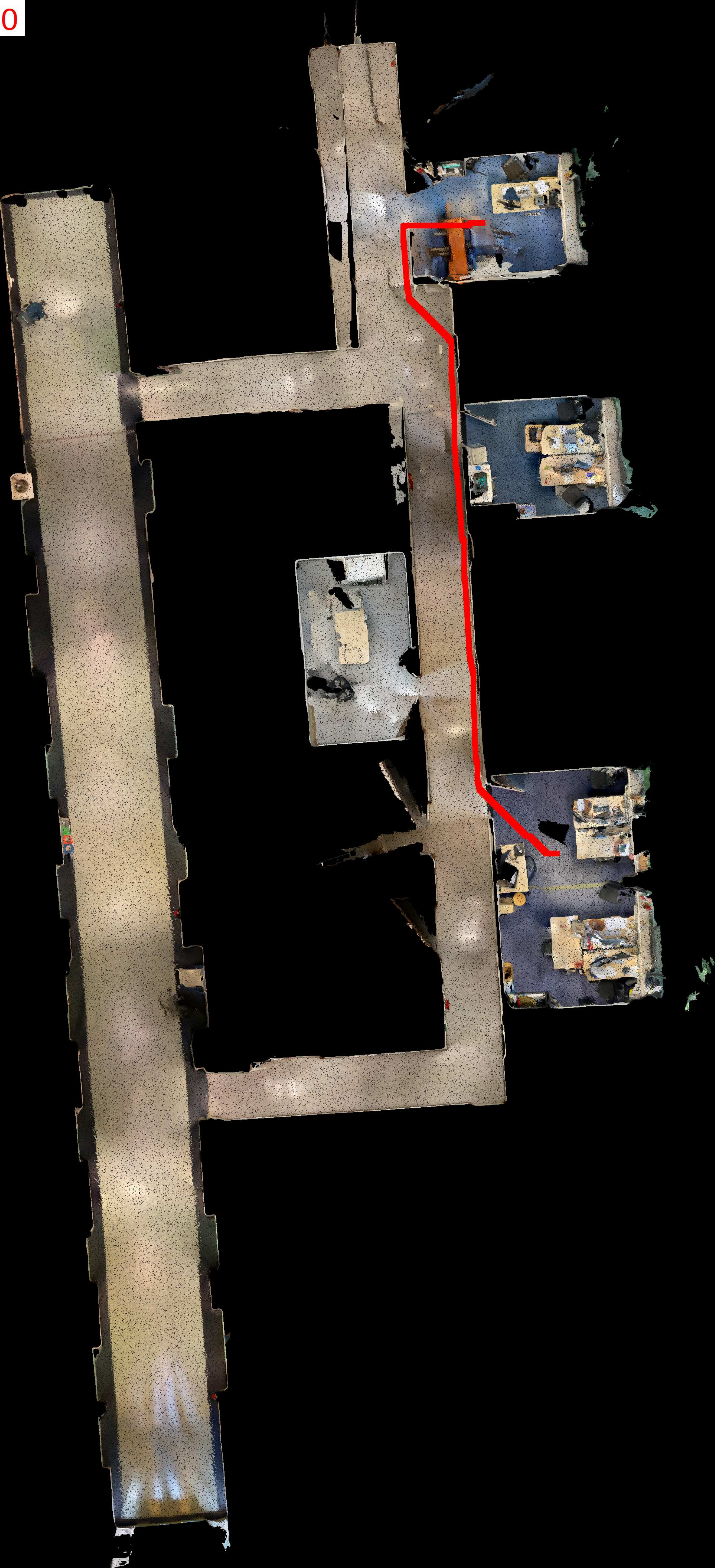}
}
\subfloat[path\_1]
{
    \includegraphics[width=2.585cm]{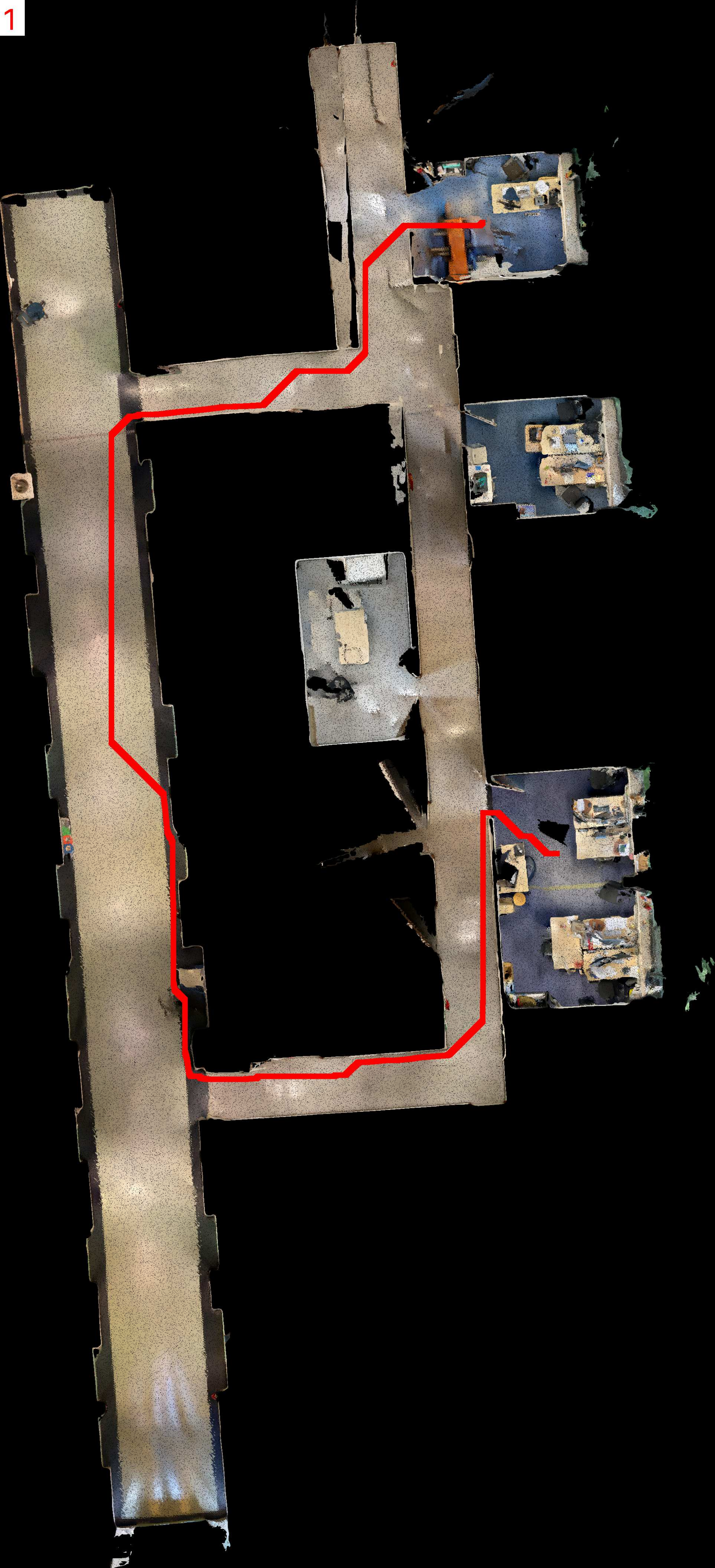}
}
\subfloat[map]
{
    \includegraphics[width=2.745cm]{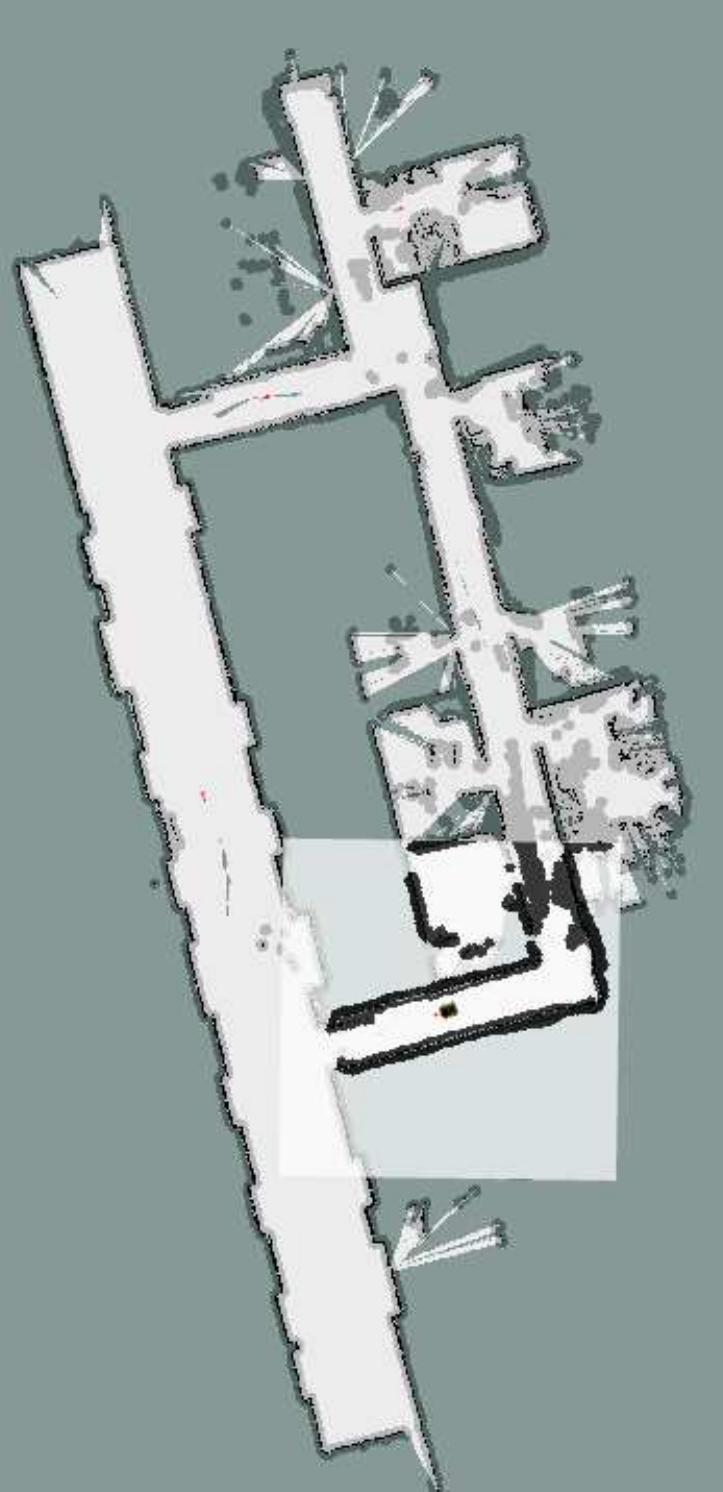}
}
\caption{Privacy-aware navigation in a real office scene. The red lines in (a) and (b) represent the paths searched from the topological map, and (c) shows the robot navigating with the cost map in ROS RVIZ.}
\label{fig:realmap}
\vspace{-0.4cm}
\end{figure}

We utilized ROS and the Jackal Robot hardware platform with Ouster lidar to implement and test our framework in the real world. To minimize the gap between the S3DIS dataset and real-world scenarios, we formulated the real scene with the same representation. Specifically, we scanned an office scene with three office rooms, a conference room, and a few corridors using the 3D Scanner on iPad Pro and obtained the 3D point cloud of the scene as the scene model, as shown in Fig. \ref{fig:realscene}. The top-view map and traversability map can be derived from the point cloud. Each room in the traversability map was manually marked with a navigational point as a node, and each pair of neighboring nodes was connected, forming the topological map of the scene. Given the navigational instruction: \emph{send a sensitive file from office\_1 to office\_3}, all potential paths were searched from the topological map and planned in the metric maps. As shown in \ref{fig:realmap} (a-b), path 0 is the shortest but passes through a more crowded office area. Path 1, although it takes a longer route, minimizes exposure to densely populated office areas, ensuring better privacy and security for sensitive files.

After getting the optimal path for navigation, we align the scanned scene model with the metric maps built by the lidar on the robot, so the robot can use the path to complete navigation. In particular, we use Gmapping \cite{Grisetti2005} with Ouster lidar to build the grid map of the scene and manually align it with the traversability map. This allows us to obtain the positions of the topological nodes in the grid map, and the navigation stack \cite{Marder-Eppstein2010} is used to complete the navigation task following the topological path, as shown in Fig. \ref{fig:realmap} (c). 
Furthermore, scanning the scene in advance and aligning the scene model to the real world has an advantage in sim2real transfer, as it uses the same scene representation. This reduces the impact of noise from lidar, allowing us to deploy the model on the robot platform with minimal fine-tuning.

\section{CONCLUSIONS}

This paper presented PANav, a novel framework that leverages vision-language models to facilitate privacy-aware navigation by developing an adaptive path planner. By conducting experiments on the S3DIS dataset, we demonstrate that our approach significantly improves the privacy awareness of the navigation path in human-shared public environments for mobile robots. We also validated its effectiveness in a real-world experiment, highlighting the practical applicability of our method.
Future research should incorporate additional hidden information, such as the robot's state, environmental features, and specific navigational instructions. Extending our method to dynamic settings, such as when encountering people during navigation, would also be a valuable direction.



\bibliographystyle{ieeetr}
\bibliography{bib/library.bib}


\end{document}